\documentclass[onecolumn,11pt]{article}

\RequirePackage{fix-cm}
\usepackage{graphicx}

\usepackage{enumitem}

\usepackage{microtype}

\usepackage{times}
\usepackage{tabu}
\usepackage{courier}
\usepackage{soul,color}

\usepackage{hyperref}

\usepackage{caption}
\usepackage{subcaption}

\usepackage{amsmath}
\usepackage{amssymb}

\usepackage{xypic}

\usepackage[all,2cell]{xy}\UseAllTwocells\SilentMatrices
\usepackage{multirow}

\usepackage{color}

\usepackage{graphicx}
\usepackage{url}
\usepackage{color}
\begin{document}

\title{\textbf{UNCC Biomedical Semantic Question Answering Systems. \\
BioASQ: Task-7B, Phase-B.  \\
}}

\author{ Sai Krishna Telukuntla, Aditya Kapri \and \& Wlodek Zadrozny \\
College of Computing and Informatics (CCI),\\ UNC Charlotte NC 28223, USA \\
\{stelukun,akapri,wzadrozn\}@uncc.edu}


{}

\maketitle

%
%
%
%
%
%
\maketitle              
\begin{abstract}
In this paper, we detail our submission to the 7th year BioASQ competition. We present our approach for Task-7b, Phase B, Exact Answering Task. These Question Answering (QA) tasks include Factoid, Yes/No, List Type Question answering. Our system is based on a contextual word embedding model. 
We have used a Bidirectional Encoder Representations from
Transformers(BERT) based system, fined tuned for biomedical question answering task using BioBERT.
In the third test batch set, our system achieved the highest ‘MRR’ score for Factoid Question Answering task. Also, for List type question answering task our system achieved the highest recall score in the fourth test batch set.
 Along with our detailed approach, we present the results for our submissions, and also highlight identified downsides for our current approach and ways to improve them in our future experiments.

\end{abstract}
\section{Introduction}

BioASQ \footnote{ \url{http://BioASQ.org/participate/challenges }} is a biomedical document classification, document retrieval, and question answering competition, currently in its seventh year. We provide an overview
of our submissions to semantic question answering task (7b, Phase B) of BioASQ 7 (except for 'ideal answer' test, in which we did not participate this year). In this task systems are provided with biomedical questions and are required to submit ideal
and exact answers to those questions. We have used BioBERT \cite{Lee2019BioBERTAP} based system , see also Bidirectional Encoder Representations from Transformers(BERT) \cite{Devlin2018BERTPO}, and we fine tuned it 
for the biomedical question answering task. Our system scored near the top for factoid questions for all the batches of the challenge. More specifially, in the third test batch set, our system achieved highest ‘MRR’ score for Factoid Question Answering task. Also, for List-type question answering task our system achieved highest recall score in the fourth test batch set. Along with our detailed approach, we present the results for our submissions and also highlight identified downsides for our current approach and ways to improve them in our future experiments.  In last test batch results we placed 4th for List-type questions and 3rd for Factoid-type questions.) \\

The QA task is organized in two phases. Phase A deals with retrieval of the relevant document, snippets, concepts, and RDF triples, and phase B deals with exact
and ideal answer generations (which is a paragraph size summary of snippets). Exact answer generation is required for factoid, list, and yes/no type question.

BioASQ organizers provide the training and testing data. The training data consists of questions, gold standard documents, snippets, concepts, and ideal
answers (which we did not use in this paper, but we used last year \cite{bhandwaldar2018uncc}). The test data is split between phases A and B. The Phase A dataset consists of the questions, unique ids, question types. The Phase B dataset
consists of the questions, golden standard documents, snippets, unique ids and question types. Exact answers for factoid type questions are evaluated using strict
accuracy (the top answer), lenient accuracy (the top 5 answers), and MRR (Mean Reciprocal Rank) which takes into account the ranks of returned
answers. Answers for the list type question are evaluated using precision, recall, and F-measure. %

\section{Related Work}

\subsection{BioAsq}

Sharma et al. \cite{Sharma2018BioAMATA} describe a system with two stage process for factoid and list type question answering. Their system extracts relevant entities and then runs supervised classifier to rank the entities.
Wiese et al. \cite{DBLP:conf/bionlp/WieseWN17} propose neural network based model for Factoid and List-type question answering task. The model is based on Fast QA and predicts the answer span in the passage for a given question. The model is trained on SQuAD data set and fine tuned on the BioASQ data.
Dimitriadis et al. \cite{Dimitriadis2019WordEA}  proposed two stage process for Factoid question answering task. Their system uses general purpose tools such as Metamap, BeCas to identify candidate sentences. These candidate sentences are represented in the form of features, and are then ranked by the binary classifier. Classifier is trained on candidate sentences extracted from relevant questions, snippets and correct answers from BioASQ challenge.
For factoid question answering task highest ‘MRR’ achieved in the 6th edition of  BioASQ competition is ‘0.4325’. 
Our system is a neural network model based on contextual word embeddings \cite{Devlin2018BERTPO} and achieved a ‘MRR’ score ‘0.6103’ in one of the test batches for Factoid Question Answering task.

\subsection{A minimum background on BERT}

\begin{figure}[!h]
\includegraphics[width=\textwidth]{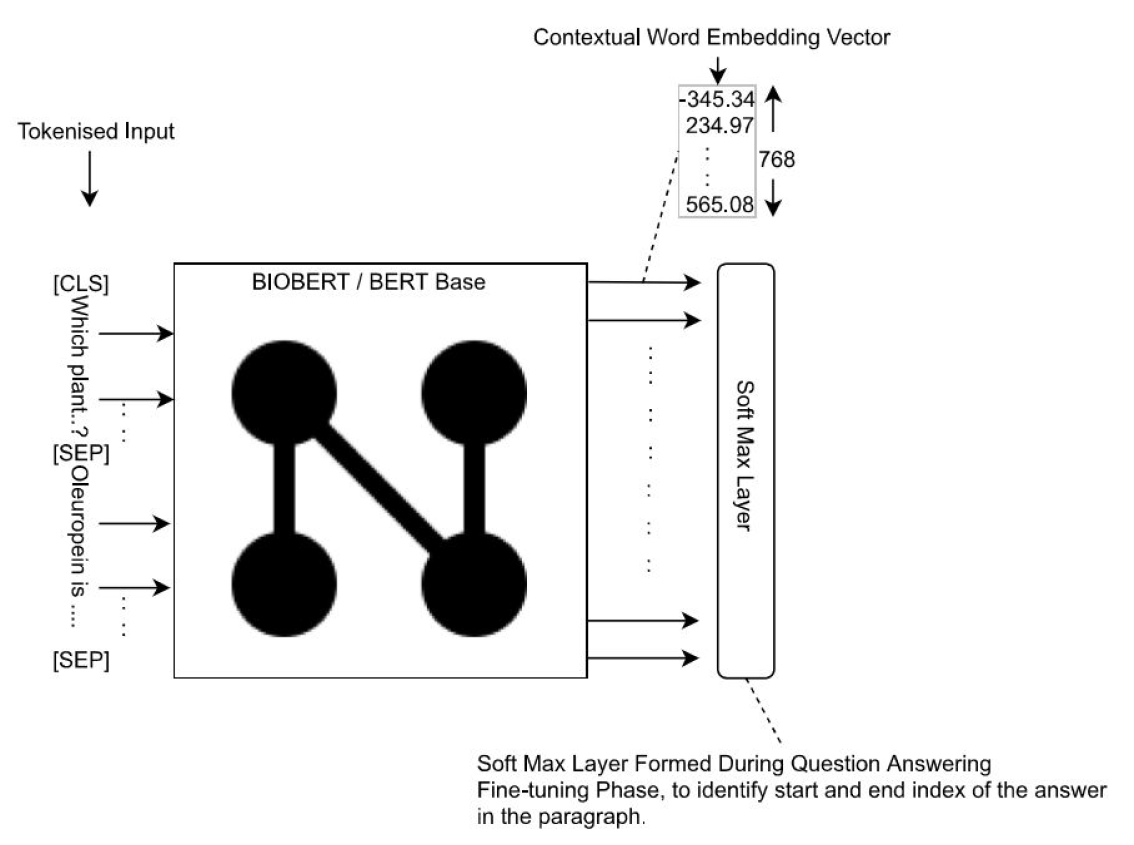}
\caption{BioBERT fine tuned for question answering task} \label{Fig1}
\end{figure}

BERT stands for "Bidirectional Encoder Representations from
Transformers" \cite{Devlin2018BERTPO} is a contextual word embedding model. Given a sentence as an input, contextual embedding for the words are returned. The BERT model was designed  so it can be fine tuned for 11 different tasks \cite{Devlin2018BERTPO}, including question answering tasks. For a question answering task, question and paragraph (context) are given as an input. A BERT standard is that question text  and paragraph text are separated by a separator [Sep]. BERT question-answering fine tuning involves adding softmax layer. Softmax layer takes contextual word embeddings from BERT as input and learns to identity answer span  present in the paragraph (context). This process is represented in Figure \ref{Fig1}.
  
BERT was originally trained to perform tasks such as language model creation using masked words  and next-sentence-prediction. In other words BERT weights are learned such that context is used in building the representation of the word, not just as a loss function to help learn a context-independent representation. 
 For detailed understanding of BERT Architecture, please refer to the original BERT paper \cite{Devlin2018BERTPO}.  
\subsubsection{Comparison of Word Embeddings and Contextual Word Embeddings}

A ‘word embedding’ is a learned representation. It is represented in the form of vector where words that have the same meaning have a similar vector representation. Consider a word embedding model 'word2vec' \cite{mikolov2013efficient} trained on a corpus. Word embeddings generated from the model are context independent that is, word embeddings are returned regardless of where the words appear in a sentence and regardless of e.g. the sentiment of the sentence. However, contextual word embedding models like BERT also takes context of the word into consideration.

\subsection{Comparison of BERT and Bio-BERT}

‘BERT’ and BioBERT are very similar in terms of architecture. Difference is that ‘BERT’ is pretrained on Wikipedia articles, whereas BioBERT version used in our experiments is pretrained on Wikipedia, PMC and PubMed articles. Therefore BioBERT model is expected to perform well with biomedical text, in terms of generating contextual word embeddings. 

BioBERT model used in our experiments is based on BERT-Base Architecture;  BERT-Base has 12 transformer Layers where as BERT-Large has 24 transformer layers. Moreover contextual word embedding vector size is 768 for BERT-Base and more for BERT-large. According to \cite{Devlin2018BERTPO} Bert-Large, fine-tuned on SQuAD 1.1 question answering data \cite{Rajpurkar2016SQuAD10} can achieve F1 Score of 90.9 for Question Answering task where as BERT-Base Fine-tuned on the same SQuAD question answering \cite{Rajpurkar2016SQuAD10} data could achieve F1 score of 88.5. One downside of the current version BioBERT is that word-piece vocabulary  
\footnote{ vocab.txt and all other software and data is available in our GitHub Repo \url{https://github.com/telukuntla/BioMedicalQuestionAnswering_UNCC} }
is the same as that of original BERT Model, as a result word-piece vocabulary does not include biomedical jargon. Lee et al. \cite{Lee2019BioBERTAP} created BioBERT, using the same pre-trained BERT released by Google, and hence in the word-piece vocabulary (vocab.txt), as a result biomedical jargon is not included in word-piece vocabulary. Modifying word-piece vocabulary (vocab.txt) at this stage would loose original compatibility with ‘BERT’, hence it is left unmodified. 

In our future work we would like to build pre-trained ‘BERT’ model from scratch. We would pretrain the model with biomedical corpus (PubMed, ‘PMC’) and  Wikipedia. Doing so would give us scope to create word piece vocabulary to include biomedical jargon and there are chances of model performing better with biomedical jargon being included in the word piece vocabulary. We will consider this scenario in the future, or wait for the next version of BioBERT.

\section{Experiments: Factoid Question Answering Task}

For Factoid Question Answering task, we fine tuned BioBERT \cite{Lee2019BioBERTAP} with question answering data and added new features. Fig. \ref{Fig1} shows the architecture of BioBERT fine tuned for question answering tasks:
Input to BioBERT is word tokenized embeddings for question and the paragraph (Context). As per the ‘BERT’ \cite{Devlin2018BERTPO} standards, tokens ‘[CLS]’ and ‘[SEP]’ are appended to the tokenized input as illustrated in the figure. The resulting model has a softmax layer formed for predicting answer span indices in the given paragraph (Context). On test data, the fine tuned model generates $n$-best predictions for each question. For a question, $n$-best corresponds that $n$ answers are returned as possible answers in the decreasing order of confidence. Variable $n$ is configurable. In our paper, any further mentions of ‘answer returned by the model’ correspond to the top answer returned by the model.

\subsection{Setup}

BioASQ provides the training data. This data is based on previous BioASQ competitions. Train data we have considered is aggregate of all train data sets till the 5th version of BioASQ competition. We cleaned the data, that is, question-answering data without answers are removed and left with a total count of ‘530’ question answers. The data is split into train and test data in the ratio of 94 to 6; that is, count of '495' for training and '35' for testing. 

The original data format is converted to the BERT/BioBERT format, where BioBERT expects ‘start{\_}index’ of the actual answer. The ‘start{\_}index corresponds to the index of the answer text present in the paragraph/ Context. For finding ‘start\_index’ we used built-in python function find(). The function returns the lowest index of the actual answer present in the context(paragraph). If the answer is not found ‘-1’ is returned as the index. The efficient way of finding start{\_}index is, if the paragraph (Context) has multiple instances of answer text, then ‘start{\_}index’ of the answer should be that instance of answer text whose context actually matches with what’s been asked in the question.

\noindent
\textbf{Example} (Question, Answer and Paragraph from \cite{bioasq2015a}):

\noindent
\textit{Question:} Which drug should be used as an antidote in benzodiazepine overdose? 

\noindent
\textit{Answer:} 'Flumazenil'

\noindent
\textit{Paragraph(context):}
\begin{quote}

"Flumazenil use in benzodiazepine overdose in the UK: a retrospective survey of NPIS data. OBJECTIVE: Benzodiazepine (BZD) overdose (OD) continues to cause significant morbidity and mortality in the UK. \textbf{Flumazenil} is an effective antidote but there is a risk of seizures, particularly in those who have co-ingested tricyclic antidepressants. A study was undertaken to examine the frequency of use, safety and efficacy of flumazenil in the management of BZD OD in the UK. METHODS: A 2-year retrospective cohort study was performed of all enquiries to the UK National Poisons Information Service involving BZD OD. RESULTS: Flumazenil was administered to 80 patients in 4504 BZD-related enquiries, 68 of whom did not have ventilatory failure or had recognised contraindications to flumazenil. Factors associated with flumazenil use were increased age, severe poisoning and ventilatory failure. Co-ingestion of tricyclic antidepressants and chronic obstructive pulmonary disease did not influence flumazenil administration. Seizure frequency in patients not treated with flumazenil was 0.3\%".

\end{quote}
 
Actual answer is 'Flumazenil', but there are multiple instances of word 'Flu-mazenil'. Efficient way to identify the start-index for 'Flumazenil'(answer) is to find that particular instance of the word 'Flumazenil' which matches the context for the question. In the above example 'Flumazenil' highlighted in bold is the actual instance that matches question's context. Unfortunately, we could not identify readily available tools that can achieve this goal. In our future work, we look forward to handling these scenarios effectively. 

Note: The creators of 'SQuAD' \cite{Rajpurkar2016SQuAD10}   have handled the task of identifying answer's start{\_}index effectively. But 'SQuAD' data set is much more general and does not include biomedical question answering data.

\subsection{Training and error analysis}
During our training with the BioASQ data, learning rate is set to 3e-5, as mentioned in the BioBERT paper \cite{Lee2019BioBERTAP}. We started training the model with 495 available train data and 35 test data by setting number of epochs to 50. After training with these hyper-parameters training accuracy(exact match) was 99.3\%(overfitting) and testing accuracy is only 4\%. In the next iteration we reduced the number of epochs to 25 then training accuracy is reduced to 98.5\% and test accuracy moved to 5\%. We further reduced number of epochs to 15, and the resulting training accuracy was 70\% and test accuracy 15\%. In the next iteration set number of epochs to 12 and achieved train accuracy of 57.7\% and test accuracy 23.3\%. Repeated the experiment with 11 epochs and found training accuracy to be 57.7\% and test accuracy to be same 22\%. In the next iteration we set number of epochs to '9' and found training accuracy of 48\% and test accuracy of 15\%. Hence optimum number of epochs is taken as 12 epochs. 

During our error analysis we found that on test data, model tends to return text in the beginning of the context(paragraph) as the answer. On analysing train data, we found that there are '120'(out of '495') question answering data instances having start{\_}index:0, meaning 120(~25\%) question answering data has first word(s) in the context(paragraph) as the answer. We removed 70\% of those instances in order to make train data more balanced. 
In the new train data set we are left with '411' question answering data instances. This time we got the highest test accuracy of 26\% at 11 epochs. We have submitted our results for BioASQ test batch-2, got strict accuracy of 32\% and our system stood in 2nd place. Initially, hyper-parameter- 'batch size' is set to ‘400’. Later it is tuned to '32'. Although accuracy(exact answer match) remained at 26\%, model generated concise and better answers at batch size ‘32’, that is wrong answers are close to the expected answer in good number of cases.

\noindent
\textbf{Example.}(from \cite{bioasq2015a})

\noindent
\textit{Question:} Which mutated gene causes Chediak Higashi Syndrome? 

\noindent
\textit{Exact Answer:} ‘lysosomal trafficking regulator gene’.

The answer returned by a model trained at ‘400’ batch size is \textit{‘Autosomal-recessive complicated spastic paraplegia with a novel lysosomal trafficking regulator’}, and from the one trained at ‘32’ batch size is \textit{‘lysosomal trafficking regulator’}.

In further experiments, we have fine tuned the BioBERT model with both ‘SQuAD’ dataset (version 2.0) and BioAsq train data. For training on ‘SQuAD’, hyper parameters- Learning rate and number of epochs are set to ‘3e-3’ and ‘3’ respectively as mentioned in the paper \cite{Devlin2018BERTPO}. 
Test accuracy of the model boosted to 44\%. In one more experiment we trained model only on ‘SQuAD’ dataset, this time test accuracy of the model moved to 47\%. The reason model did not perform up to the mark when trained with ‘SQuAD’ alongside BioASQ data could be that in formatted BioASQ data, start{\_}index for the answer is not accurate, and affected the overall accuracy.

\section{Our Systems and Their Performance on Factoid Questions}

We have experimented with several systems and their variations, e.g. created by training with specific additional features (see next subsection).
Here is their list and short descriptions. Unfortunately we did not pay attention to naming, and the systems evolved between test batches, so the overall picture can only be understood by looking at the details.

When we started the experiments our objective was to see whether BioBERT and entailment-based techniques can provide value for in the context of biomedical question answering. The answer to both questions was a yes, qualified by many examples clearly showing the limitations of both methods. Therefore we tried to address some of these limitations using feature engineering with mixed results: some clear errors got corrected and new errors got introduced, without overall improvement but convincing us that in future experiments it might be worth trying feature engineering again especially if more training data were available.

Overall we experimented with several approaches with the following aspects of the systems changing between batches, that is being absent or present: 
\begin{itemize}
\item[*] training on BioAsq data vs. training on SQuAD 
\item[*] using the BioAsq snippets for context vs. using the documents from the provided URLs for context 
\item[*] adding or not the LAT, i.e. lexical answer type, feature \ (see \cite{lally2012question}, \cite{brown2012system} and an explanation in the subsection just below).
\end{itemize}
For Yes/No questions (only) we experimented with the entailment methods.

We will discuss the performance of these models below and in Section 6. But before we do that, let us discuss a feature engineering experiment which eventually produced mixed results, but where we feel it is potentially useful in future experiments.

\subsection{LAT Feature considered and its impact (slightly negative)}
During error analysis we found that for some cases, answer being returned by the model is far away from what it is being asked in the Question.  

\noindent
\textbf{Example:} (from \cite{bioasq2015a})

\noindent
\textit{Question:} Hy's law measures failure of which organ?

\noindent
\textit{Actual Answer: }‘Liver’. 

The answer returned by one of our models was \textit{‘alanine aminotransferase’}, which is an enzyme. The model returns an enzyme, when the question asked for the organ name. To address this type of errors, we decided to try the concepts of
‘Lexical Answer Type’ (LAT) and Focus Word, which was used in IBM Watson, see \cite{Ferrucci2010BuildingWA} for overview; \cite{brown2012system} for technical details, and \cite{lally2012question} for details on question analysis.  
In an example given in the last source we read:
\begin{quote}
\textit{POETS \& POETRY: He was a bank clerk in the
Yukon before he published "Songs of a Sourdough"
in 1907.}

The focus is the part of the question that is a reference
to the answer. In the example above, the focus is "he". 

LATs are terms in the question that indicate what type
of entity is being asked for. 

The headword of the focus is
generally a LAT, but questions often contain additional
LATs, and in the Jeopardy! domain, categories are an
additional source of LATs. 
 
(...)
In the example, LATs are "he",
"clerk", and "poet".
\end{quote}
\noindent
For example in the question \textit{ "Which plant does oleuropein originate from?"} (\cite{bioasq2015a}).
The LAT here is ‘plant’. For the BioAsq task we did not need to explicitly distinguish between the focus and the LAT concepts.
In this example, the expectation is that answer returned by the model is a plant. Thus it is conceivable that the cosine distance between contextual embedding of word 'plant' in the question and contextual embedding for the answer present in the paragraph(context) is comparatively low. As a result model learns to adjust its weights during training phase and returns answers with low cosine distance with the LAT.

We used Stanford CoreNLP \cite{Manning2014TheSC} library to write rules for extracting lexical answer type present in the question, both 'parts of speech'(POS) and dependency parsing functionality was used. We incorporated the Lexical Answer Type into one of our systems, UNCC{\_QA1} in Batch 4. 
This system underperformed our system FACTOIDS by about 3\% in the MRR measure, but corrected errors such as in the example above.

\subsubsection{Assumptions and rules for deriving lexical answer type.} 

\begin{figure}[ht!]
\begin{center}
\includegraphics[scale=0.85]{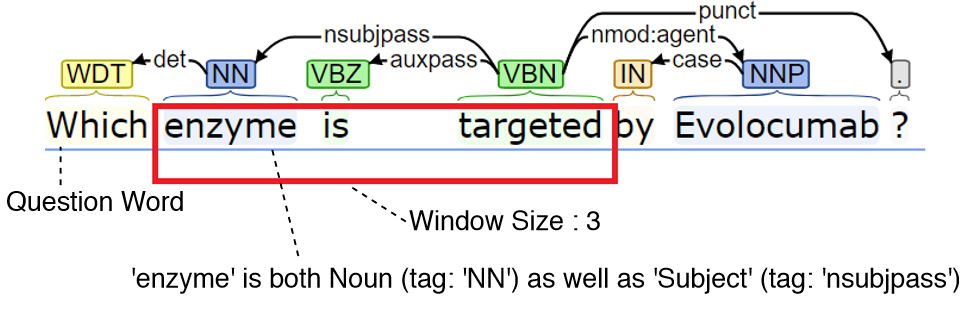}

\bigskip
\includegraphics[scale=0.9]{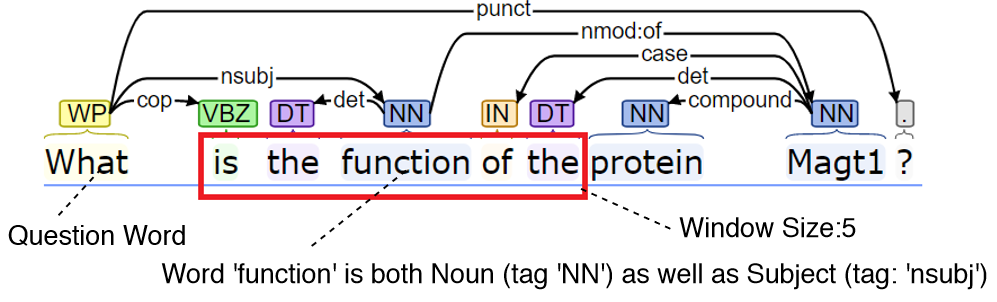}
\caption{A simple way of finding the lexical answer types, LATs, of factoid questions: using POS tags to find the question word (e.g. 'which'), and a dependency parse to find the LAT within the window of 3  words. If a noun is not found near the "Wh-" word, we iterate looking for it, as in the second panel.} 
\label{dpars}
\end{center}
\end{figure}

\noindent 
There are different question types: ‘Which’, ‘What’, ‘When’, ‘How’ etc. Each type of question is being handled differently and there are commonalities among the rules written for different question types. 
Question words are identified  through parts of speech tags: 'WDT', 'WRB' ,'WP'. We assumed that LAT is a ‘Noun’ and follows the question word. Often it was also a subject (\textsc{nsubj}). This process is illustrated in Fig.\ref{dpars}. 

LAT computation was governed by a few simple rules, e.g. when a question has multiple words that are 'Subjects’ (and ‘Noun’), a word that is in proximity to the question word is considered as ‘LAT’.  These rules are different for each "Wh" word. 

  
Namely, 
when the word immediately following the question word is a \texttt{Noun}, window size is set to ‘3’. The window size ‘3’ means we iterate through the next ‘3’ words to check if any of the word is both  \texttt{Noun} and \texttt{Subject}, If so, such word is considered the ‘LAT’; else the word that is present very next to the question word is considered as the ‘LAT’. 

For questions with words ‘Which’ , ‘What’, ‘When’; a \textit{Noun} immediately following the question word is very often the LAT, e.g. 'enzyme' in   
\textit{Which enzyme is targeted  by Evolocumab?}. When the word immediately following the question word is not a \texttt{Noun}, e.g. in \textit{What is the function of the protein Magt1?} the window size is set to ‘5’, and we iterate through the next ‘5’ words (if present) and search for the word that is both \texttt{Noun} and \texttt{Subject}. If present, the word is considered as the ‘LAT’; else, the \texttt{Noun} in close proximity to the question word and following it is returned as the ‘LAT’. 

For questions with question words: ‘When’, ‘Who’, ‘Why’, the ’LAT’ is a question word itself. 
For the word ‘How', e.g. in \textit{How many selenoproteins are encoded in the human genome?}, we look at the adjective and if we find one, we take it to be the LAT, otherwise the word 'How' is considered as the ‘LAT’. \\
 

Perhaps because of using only very simple rules, the accuracy for ‘LAT’ derivation is 75\%; that is, in the remaining 25\% of the cases the LAT word is identified incorrectly. Worth noting is that the overall performance the system that used LATs was slightly inferior to the system without LATs, but the types of errors changed.
The training used BioBERT with the LAT feature as part of the input string. 
The errors it introduces usually involve 
 finding the wrong element of the correct type e.g. wrong enzyme when two similar enzymes are described in the text, or 'neuron' when asked about a type of cell with a certain function, when the answer calls for a different cell category, adipocytes, and both are mentioned in the text.
 We feel with more data and additional tuning or perhaps using an ensemble model, we might be able to keep the correct answers, and improve the results on the confusing examples like the one mentioned above. Therefore if we improve our ‘LAT’ derivation logic, or have larger datasets, then perhaps the neural network techniques they will yield better results.

\begin{figure}[ht!]
\begin{center}
\includegraphics[width=\textwidth]{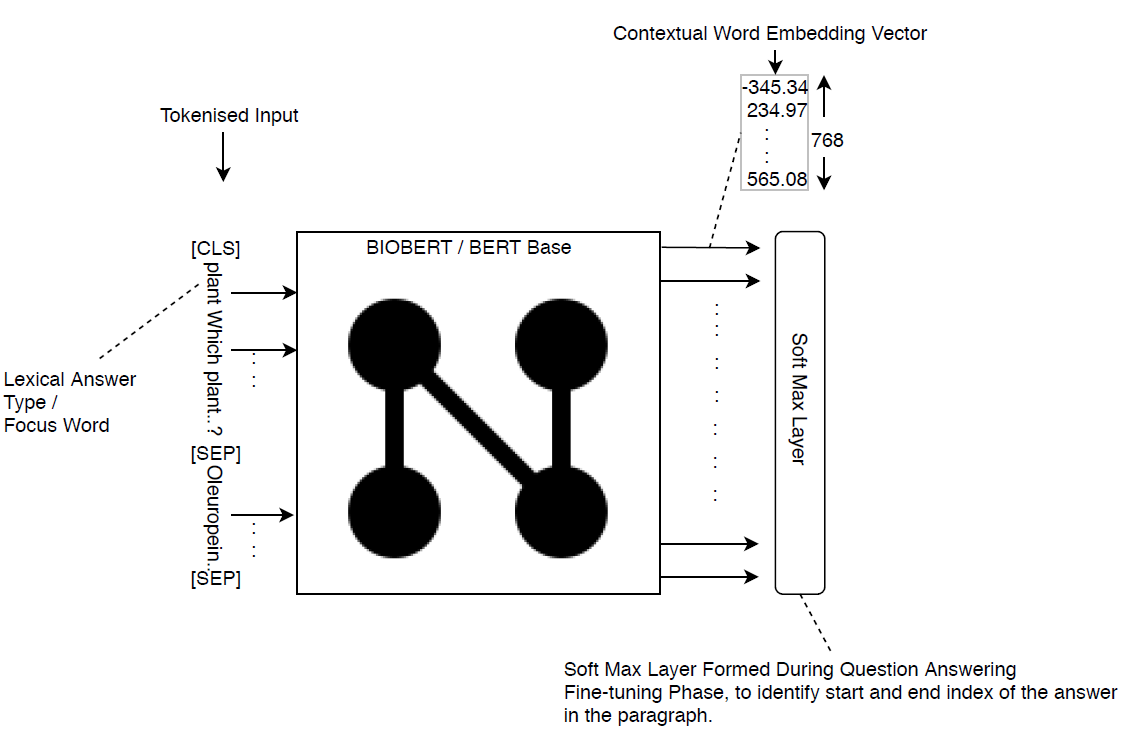}
\caption{ 
An example of a using BioBERT with additional features: Contextual word  embedding for Lexical Answer Type (LAT) given as feature along with the actual contextual embeddings for the words in question and the paragraph. This change produced mixed results and no overall improvement.} 
\label{fig:fig7}
\end{center}
\end{figure}

\subsection{Impact of Training using BioAsq data (slightly negative)}

Training on BioAsq data in our entry in Batch 1 and Batch 2  under the name {QA1} 
showed it might lead to overfitting. This happened both with (Batch 2) and without (Batch 1) hyperparameters tuning: abysmal 18\% MRR in Batch 1, and slighly better one, 40\% in Batch 2
(although in Batch 2 it was overall the second best result in MRR but 16\% lower than the highest score).

In Batch 3 (only), our UNCC{\_}QA3 system was fine tuned on BioAsq and SQuAD 2.0 \cite{Rajpurkar2016SQuAD10}, and for data preprocessing Context  paragraph is generated from relevant snippets provided in the test data. This system underperformed, by about 2\% in MRR, our other entry UNCC{\_}QA1, which was also an overall category winner for this batch. The latter was also trained on SQuAD, but not on BioAsq. We suspect that the reason could be the simplistic nature of the \textit{find()} function described in Section 3.1. So, this could be an area where a better algorithm for finding the best occurrence of an entity could improve performance.

\subsection{Impact of Using Context from URLs (negative)}

In some experiments, for context \textit{in testing}, we used documents for which URL pointers are provided in BioAsq. However, our system UNCC{\_}QA3 underperformed our other system tested only on the provided snippets. 

In Batch 5 the underperformance was about 6\%  of MRR, compared to our best system UNCC{\_}QA1, and by 9\% to the top performer.

\section{Performance on Yes/No and List questions}

Our work focused on Factoid questions. But we also have done experiments on List-type and Yes/No questions.

\subsection{Entailment improves Yes/No accuracy}

We started by answering always YES (in batch 2 and 3) to get the baseline performance. For batch 4 we used entailment.
Our algorithm was very simple: Given a question we iterate through the candidate sentences and try to find any candidate sentence is contradicting the question (with confidence over 50\%), if so 'No' is returned as answer, else 'Yes' is returned. 
In batch 4 this strategy produced better than the BioAsq baseline performance, and compared to our other systems, the use of entailment increased the performance by about 13\% (macro F1 score). We used 'AllenNlp' \cite{Gardner2018AllenNLPAD} entailment library to find entailment of the candidate sentences with question.

\subsection{For List-type the URLs have negative impact}

Overall, we followed the similar strategy that's been followed for Factoid Question Answering task.
We started our experiment with batch 2, where we submitted  20 best answers (with context from snippets). 
Starting with batch 3,  we performed post processing: once models generate answer predictions (n-best predictions), we do post-processing on the predicted answers. In test batch 4, our system (called \textsc{FACTOIDS}) achieved  highest recall score of ‘0.7033’ but low precision of 0.1119, leaving open the question of how could we have better balanced the two measures.

In the post-processing phase, we take the top ‘20’ (batch 3) and top 5 (batch 4 and 5),  predicted answers, tokenize them using common separators: 'comma' , 'and', 'also', 'as well as'. Tokens with characters count more than ‘100’ are eliminated and rest of the tokens are added to the list of possible answers. BioASQ evaluation mechanism does not consider snippets with more than ‘100’ characters as a valid answer. Considering lengthy snippets in to the list of answers would reduce the mean precision score. As a final step, duplicate snippets in the answer pool are removed. For example, consider these top 3 answers predicted by the system (before post-processing):

\begin{verbatim}
        {
            "text": "dendritic cells", 
            "probability": 0.7554540733426441, 
            "start_logit": 8.466046333312988, 
            "end_logit": 9.536355018615723
        }, 
        {
           "text": "neutrophils, macrophages and  
                    distinct subtypes of dendritic cells", 
           "probability": 0.13806867348304214, 
            "start_logit": 6.766478538513184, 
            "end_logit": 9.536355018615723
        },      
        {
         "text": "macrophages and distinct subtypes of dendritic", 
         "probability": 0.013973475271178242, 
            "start_logit": 6.766478538513184, 
            "end_logit": 7.24576473236084
        }, 
\end{verbatim}
\noindent		
After execution of post-processing heuristics, the list of answers returned is as follows:

\begin{verbatim}
            ["dendritic cells"],
            ["neutrophils"],
            ["macrophages"],
            ["distinct subtypes of dendritic cells"]
\end{verbatim}

\section{Summary of our results}

The tables below summarize all our results. They show that the performance of our systems was mixed. The simple architectures and algorithm we used worked very well only in Batch 3. However, we feel we can built a better system based on this experience. In particular we observed both the value of contextual embeddings and of feature engineering (LAT), however we failed to combine them properly.

\subsection{Factoid questions}

\begin{table}[ht!]
\centering
\caption{\textbf{Factoid Questions.} In Batch 3 we obtained the highest score. Also the relative distance between our best system and the top performing system shrunk between Batch 4 and 5.}\label{factoid}
{
\begin{tabular}{ |c|c|c|c| } 
\hline
\textbf{System} & \textbf{Strict Accuracy} & \textbf{Lenient Accuracy} & \textbf{MRR}\\
\hline 
\multicolumn{4}{c}{Batch 1} \\
\hline
QA1 & 0.1538 & 0.2308 & 0.1761\\
\hline
Top Competitor & 0.4103 & 0.5385 & 0.4637  \\
\hline
\multicolumn{4}{c}{Batch 2} \\
\hline
QA1 & 0.36 & 0.48 & 0.4033\\
\hline
Top Competitor & 0.52 & 0.64 & 0.5667  \\
\hline
\multicolumn{4}{c}{Batch 3} \\
\hline
UNCC\_QA1 & 0.4483 & 0.5862 & \textbf{0.5115}\\
\hline 
UNCC QA2 & 0.4138 & 0.5862 & 0.4856  \\ 
\hline
UNCC\_QA3 & 0.4138 & 0.5862 & 0.4943  \\ 
\hline
Top Competitor & 0.36 & 0.48 & 0.5023\\
\hline
\multicolumn{4}{c}{Batch 4} \\
\hline
FACTOIDS & 0.5294 & 0.7353 & 0.6103\\
\hline
UNCC QA1 & 0.4706 & 0.7353 & 0.5833  \\ 
\hline
Top Competitor & 0.5882 & 0.8235 & 0.6912  \\
\hline
\multicolumn{4}{c}{Batch 5} \\
\hline
UNCC{\_}QA1 & 0.2857	 & 0.4286 & 0.3305
\\
\hline 
UNCC\_QA3 & 0.2286 & 0.3143 & 0.2643\\
\hline
 QA1 & 0.2286 & 0.3714 & 0.2938  \\ 
\hline
Top Competitor & 0.2857 & 0.5143 & 0.3638  \\
\hline
\end{tabular}
}
\end{table}
\subsubsection{Systems used in Batch 5 experiments\\}

\noindent
\noindent
\textbf{System description for ‘UNCC{\_}QA1’}: The system was finetuned on the SQuAD 2.0. For data preprocessing Context / paragraph was generated from relevant snippets provided in the test data.\\
\textbf{System description for ‘QA1’ :}
‘LAT’  feature was added and finetuned with SQuAD 2.0. For data preprocessing Context / paragraph was generated from relevant snippets provided in the test data.\\
\textbf{System description for ‘UNCC{\_}QA3’} : Fine tuning process is same as it is done for the system ‘UNCC{\_}QA1’ in test batch-5. Difference is during data preprocessing, Context/paragraph is generated from the relevant documents for which URLS are included in the test data.

\subsection{List Questions}

For List-type questions, although post processing helped in the later batches, we never managed to obtain competitive precision, although our recall was good.

\begin{table}[ht!]
\centering
\caption{\textbf{List Questions}}\label{list}
{
\begin{tabular}{ |c|c|c|c| } 
\hline
\textbf{System} & \textbf{Mean Precision} & \textbf{Recall} & \textbf{F-measure}\\
\hline
\multicolumn{4}{c}{Batch 2} \\
\hline
QA1 & 0.0471 & 0.2898 & 0.0786\\
\hline
Top Competitor & 0.5826 & 0.4839 & 0.4732  \\
\hline
\multicolumn{4}{c}{Batch 3} \\
\hline
UNCC\_QA1 & 0.0780 & 0.4711 & 0.1297\\
\hline 
Top Competitor & 0.4267 & 0.3058 & 0.3298\\
\hline
\multicolumn{4}{c}{Batch 4} \\
\hline
FACTOIDS & 0.1119  & 0.7033 & 0.1893\\
\hline 
UNCC QA1 & 0.1087 & 0.6968 & 0.1846 \\ 
\hline
UNCC\_QA3 & 0.1087 & 0.6968 & 0.1846 \\ 

\hline
Top Competitor & 0.4841 & 0.5051 & 0.4604  \\
\hline
\multicolumn{4}{c}{Batch 5} \\
\hline
UNCC{\_}QA1 & 0.2051  & 0.5127 & 0.2862 \\ 
\hline
Top Competitor &  0.5653 & 0.4131 & 0.4619  \\
\hline
\end{tabular}
}
\end{table}

\subsection{Yes/No questions}
The only thing worth remembering from our performance is that using entailment can have a measurable impact (at least with respect to a weak baseline). The results (weak) are in Table 3. 
\begin{table}[h!]
\centering
\caption{\textbf{Yes/No Questions}}\label{yes/no}
{
\begin{tabular}{ |c|c|c|c|c| } 
\hline
\textbf{System} & \textbf{Accuracy} & \textbf{F1 Yes} & \textbf{F1 No} & \textbf{Macro F1}\\
\hline 
\multicolumn{5}{c}{Batch 1} \\
\hline
QA1 & 0.7931 & 0.8846 & -- &  0.4423 \\
\hline
Top Competitor & 0.8276 & 0.8980 & 0.4444 & 0.6712 \\
\hline
\multicolumn{5}{c}{Batch 2} \\
\hline
QA1 & 0.5667 & 0.7234 & -- & 0.3617\\
\hline
Top Competitor & 0.8333 & 0.8387 & 0.8276 & 0.8331 \\
\hline
\multicolumn{5}{c}{Batch 3} \\
\hline
QA1 & 0.7826 & 0.8780 & -- & 0.4390 \\
\hline 
UNCC\_QA3 & 0.7826 & 0.8780 & -- & 0.4390 \\

\hline
Top Competitor & 0.8696 & 0.9231 & 0.5714 &         0.7473 \\
\hline
\multicolumn{5}{c}{Batch 4} \\
\hline 
UNCC\_QA1 & 0.6087 & 0.7097 & 0.4000 & 0.5548 \\ 
\hline
FACTOIDS & 0.7391 & 0.8500 & -- & 0.4250\\
\hline
UNCC\_QA3 & 0.7391 & 0.8500 & -- &  0.4250 \\ 
\hline
Top Competitor & 0.8696 & 0.9143 & 0.7273 &  0.8208\\
\hline

\multicolumn{5}{c}{Batch 5} \\
\hline 
UNCC QA2 & 0.5429 & 0.7037 & -- & 0.3519 \\ 
\hline
Top Competitor & 0.8286 & 0.8500 & 0.8000 &  0.8250 \\
\hline
\end{tabular}
}
\end{table}

\section{Discussion, Future Experiments, and Conclusions}

\subsubsection{Summary: }
In contrast to 2018, when we submitted \cite{bhandwaldar2018uncc} to BioASQ a system based on extractive summarization (and scored very high in the ideal answer category), this  year we mainly targeted factoid question answering task and focused on experimenting with \textsc{BioBERT}. After these experiments we  see the promise of \textsc{BioBERT} in QA tasks, but we also see its limitations. The latter we tried to address with mixed results using feature engineering. Overall these experiments allowed us to secure a best and a second best score in different test batches. Along with Factoid-type question, we also tried ‘Yes/No’ and ‘List’-type questions, and did reasonably well with our very simple approach. 

For Yes/No the moral worth remembering is that reasoning has a potential to influence results, as evidenced by our adding the AllenNLP entailment \cite{Gardner2018AllenNLPAD} system increased its performance. 

All our data and software is available at Github, in the previously referenced URL (end of Section 2).

\subsubsection{Future experiments }

\begin{figure}[ht!]
\begin{center}
\includegraphics[width=\textwidth]{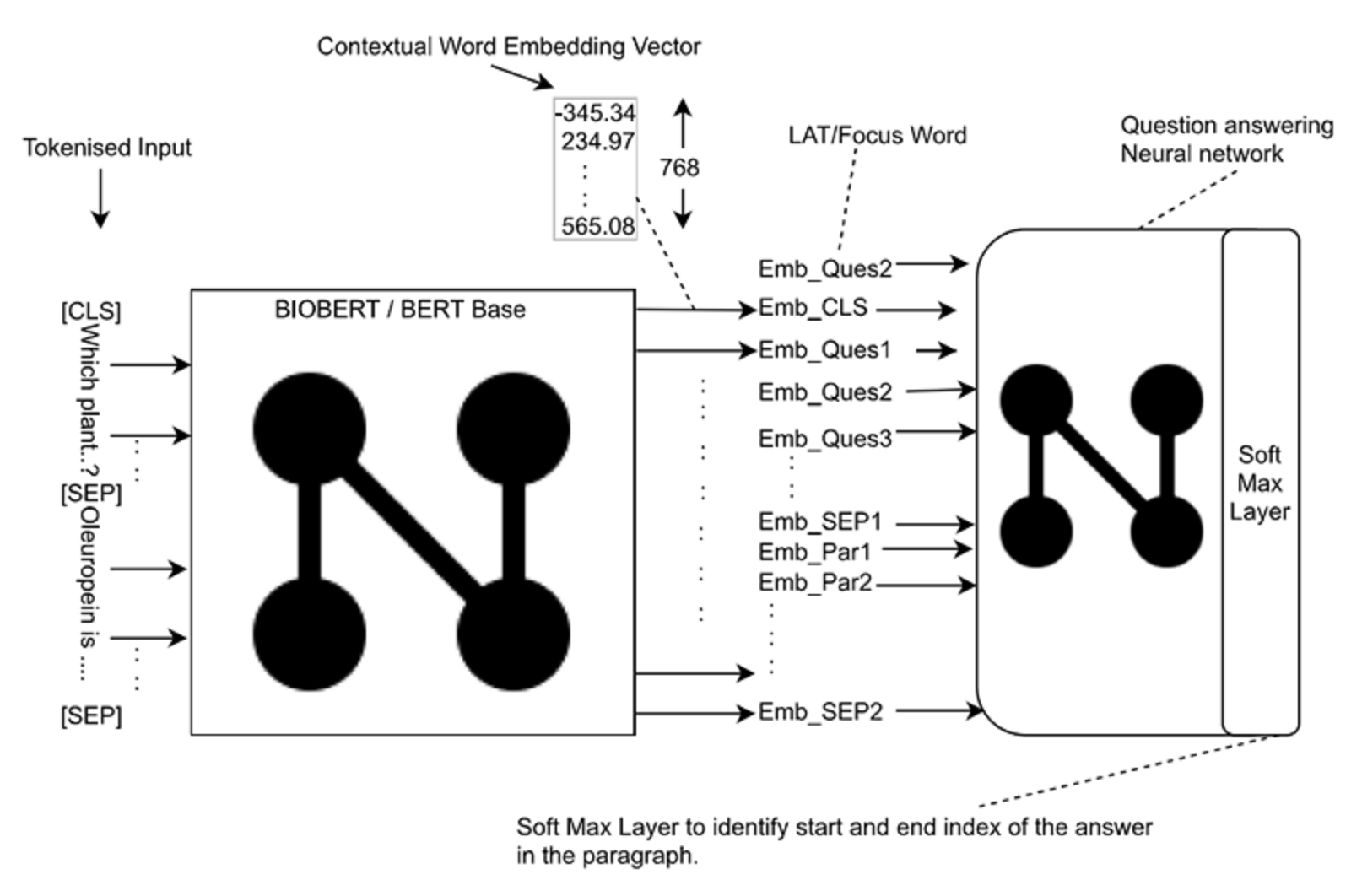}
\caption{ 
Proposed extension. In addition to using BioBERT  we propose to add a network that would train on embeddings of questions (\textit{Emb}${\_}Ques_{i}$) and paragraphs (\textit{Emb}${\_}Par_{i}$) with additional LAT features, or more broadly other semantic features, like relations that can be derived from text} 
\label{fig:fig7a}
\end{center}
\end{figure}

In the current model, we have a shallow neural network with a softmax layer for predicting answer span. Shallow networks however are not good at generalizations. In our future experiments we would like to create dense question answering neural network with a softmax layer for predicting answer span. The main idea is to get contextual word embedding for the words present in the question and paragraph (Context) and feed the contextual word embeddings retrieved from the last layer of BioBERT to the dense question answering network. The mentioned dense layered question answering neural network need to be tuned for finding right hyper parameters. An example of such architecture is shown in Fig.\ref{fig:fig7a}. 

In one more experiment, we would like to add a better version of ‘LAT’ contextual word embedding as a feature, along with the actual contextual word embeddings for question text, and Context and feed them as input to the dense question answering neural network. By this experiment, we would like to find if ‘LAT’ feature is improving overall answer prediction accuracy.
 Adding ‘LAT’ feature this way instead of feeding this word piece embedding directly to the BioBERT {(as we did in our above experiments)} would not downgrade the quality of contextual word embeddings generated form ‘BioBERT'. Quality contextual word embeddings would lead to efficient transfer learning and chances are that it would improve the model's answer prediction accuracy.
 
We also see potential for incorporating domain specific inference into the task e.g. using the MedNLI dataset \cite{romanov2018lessons}. For all types of experiments it might be worth exploring clinical BERT embeddings 
\cite{alsentzer2019publicly}, explicitly incorporating domain knowledge (e.g. \cite{lu2019incorporating}) and possibly deeper discourse representations (e.g. \cite{rao2017biomedical}).

%
%
%

\bibliographystyle{splncs04}
\bibliography{emnlp2018}

\newpage
\section{APPENDIX}

In this appendix we provide additional details about the implementations.

\subsection{Systems and their descriptions:}

We used several variants of our systems when experimenting with the BioASQ problems. In retrospect, it would be much easier to understand the changes if we adopted some mnemonic conventions in naming the systems. So, we apologize for the names that do not reflect the modifications, and necessitate this list. 

\subsubsection{Factoid Type Question Answering:}
We preprocessed the test data to convert test data to BioBERT format, We generated   Context/paragraph by either aggregating relevant snippets provided or by aggregating documents for which URLS are provided in the BioASQ test data.
\subsubsection{Batch-1:}
\subsubsection{System description for QA1:}We generated Context/paragraph by aggregating relevant snippets available in the test data and mapped it  against the question text and question id. We ignored the content present in the documents (document URLS were provided in the original test data). The model is finetuned with BioASQ data.
\subsubsection{Batch-2:}
\subsubsection{System description for QA1:}data preprocessing is done in the same way as it is done for test batch-1. Model fine tuned on BioASQ data.
\subsubsection{Batch-3:}
\subsubsection{System description for UNCC{\_}QA{\_}1:}System is finetuned on the SQuAD 2.0 [reference]. For data preprocessing Context / paragraph is generated from relevant snippets provided in the test data.
\subsubsection{System description for UNCC{\_}QA3:}System is finetuned on the SQuAD 2.0 [reference] and BioASQ dataset[].For data preprocessing Context / paragraph is generated from relevant snippets provided in the test data.
\subsubsection{System description for UNCC{\_}QA2:}Fine tuning process is same as for ‘UNCC{\_}QA{\_}1 ’. Difference is Context/paragraph is generated form from the relevant documents for which URLS are included in the test data. System ‘UNCC{\_}QA{\_}1’ got the highest ‘MRR’ score in the 3rd test batch set.
\subsubsection{Batch-4:}
\subsubsection{System description for FACTOIDS:}The System is finetuned on the SQuAD 2.0. For data preprocessing Context / paragraph is generated from relevant snippets provided in the test data.
\subsubsection{System description for UNCC{\_}QA{\_}1:}‘LAT’/ Focus word feature added and fine tuned with SQuAD 2.0 [reference]. For data preprocessing Context / paragraph is generated from relevant snippets provided in the test data.
\subsubsection{Batch-5:}
\subsubsection{System description for UNCC{\_}QA{\_}1:}The System is finetuned on the SQuAD 2.0. For data preprocessing Context / paragraph is generated from relevant snippets provided in the test data.
\subsubsection{System description for QA1:}‘LAT’/ Focus word feature added and fine tuned with SQuAD 2.0 [reference]. For data preprocessing Context / paragraph is generated from relevant snippets provided in the test data.
\subsubsection{System description for UNCC{\_}QA3:}Fine tuning process is same as it is done for the system ‘UNCC{\_}QA{\_}1’ in test batch-5. Difference is during data preprocessing, Context/paragraph is generated form from the relevant documents for which URLS are included in the test data.
\subsubsection{List Type Questions:}
We attempted List type questions starting from test batch ‘2’. Used similar approach that's been followed for Factoid Question answering task. For all the test batch sets, in the data pre processing phase Context/ paragraph is generated either by aggregating relevant snippets or by aggregating documents(URLS) provided in the BioASQ test data.

 For test batch-2, model (System: QA1) is finetuned on BioASQ data and submitted top ‘20’ answers predicted by the model as the list of answers. system ‘QA1’ achieved low F-Measure score:‘0.0786’ in the second test batch. In the further test batches for List type questions, we finetuned the model on Squad data set [reference], implemented post processing techniques (refer section 5.2) and achieved a better F-measure score: ‘0.2862’ in the final test batch set. 

In test batch-3 (Systems : ‘QA1’/’’UNCC{\_}QA{\_}1’/’UNCC{\_}QA3’/’UNCC{\_}QA2’) top 20 answers returned by the model is sent for post processing and in test batch 4 and 5 only top 5 answers are sent for post processing. System UNCC{\_}QA2(in batch 3) for List type question answering, Context is generated from documents for which URLS are provided in the BioASQ test data. for the rest of the systems (in test batch-3) for List Type question answering task snippets present in the BioaSQ test data are used  to generate context.

In test batch-4 (System : ‘FACTOIDS’/’UNCC{\_}QA{\_}1’/’UNCC{\_}QA3’’) top 5 answers returned by the model is sent for post processing. In case of system ‘FACTOIDS’ snippets in the test data were used to generate context. for systems ’UNCC{\_}QA{\_}1’ and ’UNCC{\_}QA3’ context is generated from the documents for which URLS are provided in the BioASQ test data.

In test batch-5 ( Systems: ‘QA1’/’UNCC{\_}QA{\_}1’/’UNCC{\_}QA3’/’UNCC{\_}QA2’ ) our approach is the same as that of test batch-4 where top 5 answers returned by the model is sent for post processing. for all the systems (in test batch-5) context is generated from the snippets provided in the BioASQ test data.
\subsubsection{Yes/No Type Questions:}
For the first 3 test batches, We have submitted answer ‘Yes’ to all the questions. Later, we employed ‘Sentence Entailment’ techniques(refer section 6.0) for the fourth and fifth test batch sets.
Our Systems with ‘Sentence Entailment’ approach (for ‘Yes’/ ‘No’ question answering): ‘UNCC{\_}QA{\_}1’(test batch-4),  UNCC{\_}QA3(test batch-5).

\subsection{Additional details for Yes/No Type Questions}

We used Textual Entailment in Batch 4 and 5 for ‘Yes’/‘No’ question type. The algorithm was very simple: Given a question we iterate through the candidate sentences, and look for any candidate sentences contradicting the question. If we find one 'No' is returned as answer, else 'Yes' is returned. (The confidence for contradiction was set at 50\%)
We used AllenNLP \cite{Gardner2018AllenNLPAD} entailment library to find entailment of the candidate sentences with question.

\begin{figure}[ht!]
\begin{center}
\includegraphics[scale=0.54]{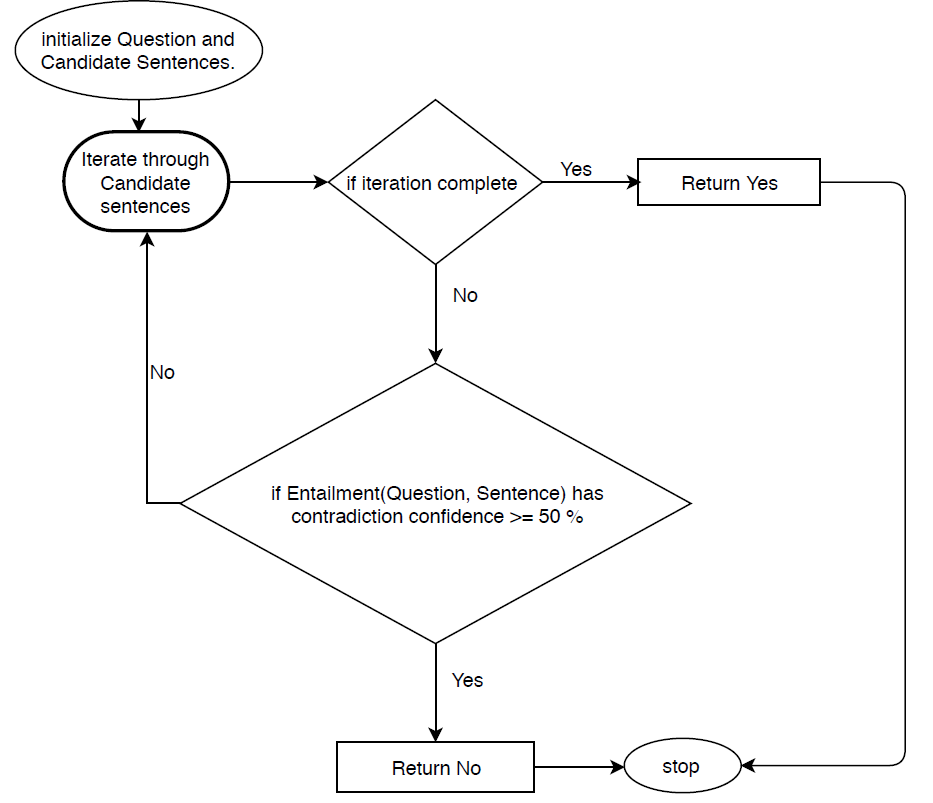}
\caption{Flow Chart for Yes/No Question answer processing} \label{Fig5}
\end{center}
\end{figure}

Flow Chart for Yes/No Question answer processing is shown in Fig.\ref{Fig5}

\newpage

\subsection{Assumptions, rules and logic flow for deriving Lexical Answer Types from  questions}
There are different question types, and we distinguished them based on the question words: ‘Which’, ‘What’, ‘When’, ‘How’ etc. Each type of question is being handled differently and there are commonalities among the rules written for different question types. 
How are question words identified? question words have parts of speech(POS): 'WDT', 'WRB', 'WP'.\\

\noindent
\textbf{Assumptions:}

1) Lexical answer type (‘LAT’) or focus word is of type \texttt{Noun} and follows the question word.

2) The LAT word is a \texttt{Subject}. 
(This clearly not always true, but we used a very simple method). 
Note: ‘StanfordNLP’ dependency parsing tag for identifying subject is  'nsubj' or 'nsubjpass'.

3) When a question has multiple words that are of type \texttt{Subject} (and \texttt{Noun}), a word that is in proximity to the question word is considered as ‘LAT’.

4) For questions with question words: ‘When’, ‘Who’, ‘Why’, the ’LAT’ is a question word itself that is, ‘When’, ‘Who’, ‘Why’ respectively.\\

\noindent
\textbf{Rules and logic flow to traverse a question:} 
The three cases below describe the logic flow of finding LATs. The figures show the grammatical structures used for this purpose.

\subsubsection{Case-1:}  Question with question word ‘How’. 

\bigskip

For questions with question word 'How', the adjective that follows the question word is considered as ‘LAT’ (need not follow immediately). If an adjective is absent, word 'How' is considered as ‘LAT’. When there are multiple words that are adjectives, a word in close proximity to the question word and follows it is returned as ‘LAT’. Note: The part of speech tag to identify adjectives is 'JJ'. For Other possible question words like ‘whose’. ‘LAT’/Focus word is question words itself.

\bigskip 
\noindent
\textbf{Example Question:} How many selenoproteins are encoded in the human genome?

\begin{figure}[!h]
\includegraphics[scale=0.33]{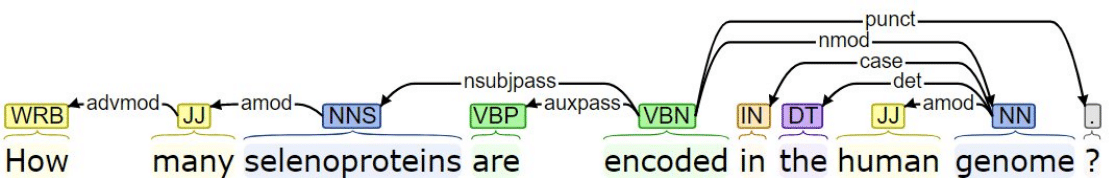}
\caption{Dependency parse visualization for the example 'how' question.} \label{img1depparsing}
\end{figure}

\subsubsection{Case-2:}  Questions with question words ‘Which’ , ‘What’ and all other possible question words; a 'Noun' immediately following the question word.

\bigskip
\noindent
\textbf{Example Question:}  Which enzyme is targeted  by Evolocumab?
\begin{figure}
\includegraphics[scale=1.1]{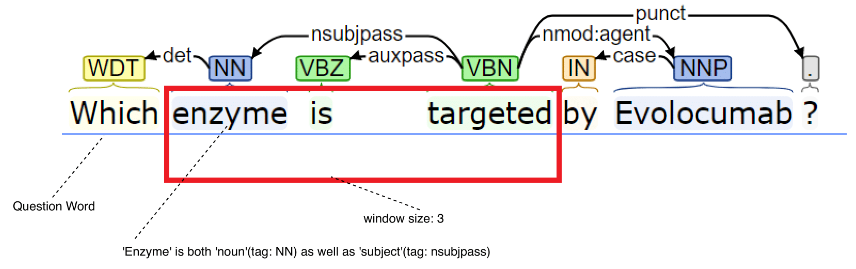}
\caption{Dependency parse visualization for the example 'which' question.} \label{img2depparsing}
\end{figure}

Here, Focus word/LAT is ‘enzyme’ which is both \texttt{Noun} and \texttt{Subject} and immediately follows the question word.

When the word immediately following the question word is  a noun, the window size is set to ‘3’. This size ‘3’ means that we iterate through the next ‘3’ words (if present) to check if any of the word is both  'Noun' and 'Subject', If so, the word is considered as ‘LAT’/Focus Word. Else the word that is present very next to the question word is considered as ‘LAT’.

\subsubsection{Case-3:}  Questions with question words ‘Which’ , ‘What’ and all other possible question words; word immediately following the question word is not a 'Noun'.

\bigskip
\noindent
\textbf{Example Question:} What is the function of the protein Magt1?

\begin{figure}
\includegraphics[scale=1]{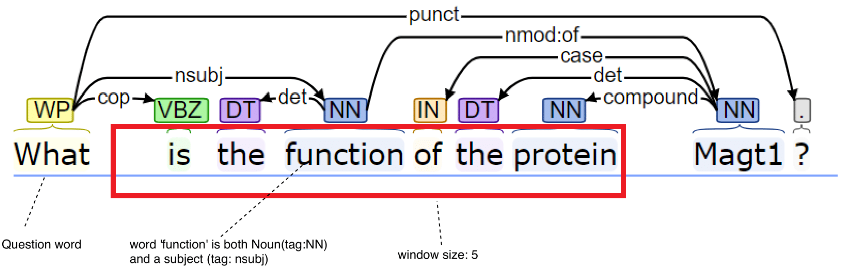}
\caption{Dependency parse visualization for the question about the function of the protein Magt1.} \label{img3depparsing}
\end{figure}

Here, Focus word/LAT is ‘function ’ which is  both \texttt{Noun} and \texttt{Subject} and does not immediately follow the question word.

When the very next word following the question word is not a \texttt{Noun}, window size is set to ‘5’. Window size ‘5’ corresponds that we iterate through the next ‘5’ words (if present) and search for the word that is both \texttt{Noun} and \texttt{Subject}. If present, the word is considered as ‘LAT’. Else, the 'Noun' close proximity to the question word and follows it is returned as ‘LAT’.

Ad we mentioned earlier, the accuracy for ‘LAT’ derivation is 75 percent. But clearly the simple logic described above can be improved, as shown in  \cite{lally2012question}, \cite{brown2012system}. Whether this in turn produces improvements in this particular task is an open question.

\subsection{Proposing Future Experiments}

In the current model, we have a shallow neural network with a softmax layer for predicting answer span. Shallow networks however are not good at generalizations. In our future experiments we would like to create dense question answering neural network with a softmax layer for predicting answer span. The main idea is to get contextual word embedding for the words present in the question and paragraph (Context) and feed the contextual word embeddings retrieved from the last layer of BioBERT to the dense question answering network. The mentioned dense layered question answering Neural network need to be tuned for finding right hyper parameters. An example of such architecture is shown in Fig.\ref{fig:fig7a}. 

In another experiment we would like to only feed contextual word embeddings for Focus word/ ‘LAT’, paragraph/ Context as input to the question answering neural network. In this experiment we would neglect all embeddings for the question text except that of Focus word/ ‘LAT’. Our assumption and idea for considering focus word and neglecting remaining words in the question is that during training phase it would make more precise for the model to identify the focus of the question and map answers against the question’s focus. To validate our assumption, we would like to take sample question answering data and find the cosine distance between contextual embedding of Focus word and that of the actual answer and verify if the cosine distance is comparatively low in most of the cases.

In one more experiment, we would like to add a better version of ‘LAT’ contextual word embedding as a feature, along with the actual contextual word embeddings for question text, and Context and feed them as input to the dense question answering neural network. By this experiment, we would like to find if ‘LAT’ feature is improving overall answer prediction accuracy. Adding ‘LAT’ feature this way instead of feeding Focus word’s word piece embedding directly (as we did in our above experiments) to the BioBERT would not downgrade the quality of contextual word embeddings generated form ‘BioBERT'. Quality contextual word embeddings would lead to efficient transfer learning and chances are that it would improve the model's answer prediction accuracy.

\noindent
\noindent

\end{document}